\pdfoutput=1

\documentclass[11pt]{article}

\usepackage[final]{coling}

\usepackage{times}
\usepackage{latexsym}
\usepackage{amsmath}
\usepackage[T1]{fontenc}

\usepackage[utf8]{inputenc}

\usepackage{microtype}

\usepackage{inconsolata}
\usepackage{natbib}
\usepackage{graphicx}
\usepackage{color}
\usepackage{caption}
\usepackage{subcaption}
\usepackage{amsmath, amssymb}
\title{Bias Vector: Mitigating Biases in Language Models with Task Arithmetic Approach}

  \author{
    \textbf{Daiki Shirafuji},
    \textbf{Makoto Takenaka}
    \textnormal{and}
    \textbf{Shinya Taguchi} \\
    \textsuperscript{}Mitsubishi Electric Corporation \\ Kamakura, Japan \\
    \texttt{\{Shirafuji.Daiki@ay,Takenaka.Makoto@bc,} \\ \texttt{Taguchi.Shinya@aj\}.MitsubishiElectric.co.jp}
  }

\begin{document}
\maketitle

\begin{abstract}
The use of language models (LMs) has increased considerably in recent years,
and the biases and stereotypes in training data that are reflected in the LM outputs
are causing social problems.
In this paper,
inspired by the task arithmetic,
we propose the ``Bias Vector'' method for the mitigation of these LM biases.
The Bias Vector method does not require manually created debiasing data.
The three main steps of our approach involve:
(1) continual training the pre-trained LMs on biased data using masked language modeling;
(2) constructing the Bias Vector as the difference between the weights of the biased LMs and those of pre-trained LMs;
and
(3) subtracting the Bias Vector from the weights of the pre-trained LMs for debiasing.
We evaluated the Bias Vector method on the SEAT across three LMs
and confirmed an average improvement of 0.177 points.
We demonstrated that the Bias Vector method does not degrade the LM performance on downstream tasks in the GLUE benchmark.
In addition, we examined the impact of scaling factors,
which control the magnitudes of Bias Vectors, with effect sizes on the SEAT
and conducted a comprehensive evaluation of our debiased LMs across both the SEAT and GLUE benchmarks.
\end{abstract}

\textcolor{red}{
\textit{Warning: This paper presents examples that can be considered discriminatory.
}}

\section{Introduction}

\begin{figure*}[t]
  \centering
  \includegraphics[width=14cm]{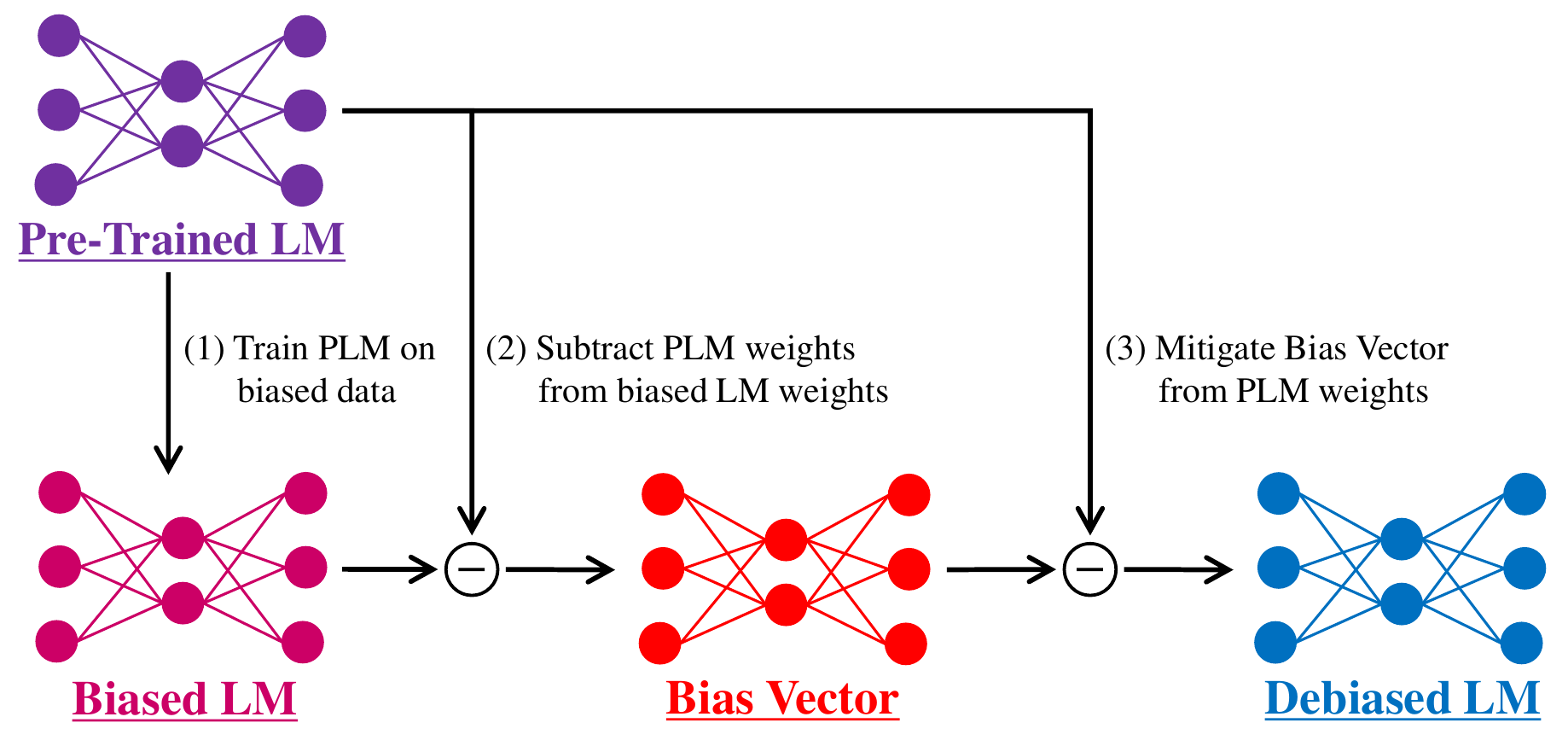}
  \caption{Overview of the Bias Vector method:
    (1) Training pre-trained LMs on biased data to create the biased models;
    (2) Subtracting pre-trained LM weights from those of the biased models
    for constructing the Bias Vectors;
    (3) Mitigating the Bias Vectors from the pre-trained LM weights for debiasing models.}
  \label{fig:overview}
\end{figure*}

As language models (LMs) have become more widely used in recent years,
the biases and stereotypes inherent in the training data for LMs
are creating social problems
\citep{liu-etal-2020-gender,kumar-etal-2023-language}.
These biases reflect the stereotypes of specific social groups
(such as those related to
\textit{race}, \textit{profession}, \textit{gender}, and \textit{religion})
\citep{NIPS2016_a486cd07,nadeem-etal-2021-stereoset}.
People tend to use racially biased stereotypical phrases
(like ``The men from afghanistan ride on \textit{camels}''),
rather than phrases that contradict stereotypes
(e.g., ``The men from afghanistan ride on \textit{skateboards}'').\footnote{The
(anti-/)stereotype examples shown are from the datasets which are publicly available on
\url{https://huggingface.co/datasets/McGill-NLP/stereoset}.}

As a consequence, LMs often make unfair predictions about certain groups,
leading to biased or stereotyped outcomes that can cause discomfort among users.
The widespread and frequent use of LMs (such as ChatGPT (GPT-3.5 / 4)
\citep{openai2022chatgpt,openai2024gpt4}),
with their biased predictions is
resulting in discrimination and inequality,
which is becoming a social problem
\citep{feng-etal-2023-pretraining}.
Hence, developing effective bias mitigation methods for LM systems is essential.

Prior to the advent of Large Language Models (LLMs),
debiasing studies primarily targeted
word embeddings
\citep{zhao-etal-2018-learning,kaneko-bollegala-2019-gender,wang-etal-2020-double}.
Models such as word2vec \citep{mikolov2013w2v}
are debiased by reshaping the word embeddings in their output representations.
However, these methods are less practical
for Transformer-based LMs, such as BERT~\citep{devlin-etal-2019-bert},
because the model parameters need to be debiased
as the required model outputs vary depending on the downstream task.

To address biases in Transformer-based LMs,
methods have been developed to reduce biases and stereotypes
by continually training of LMs with debiased datasets
\cite{zmigrod-etal-2019-counterfactual,Webster2020,dinan-etal-2020-queens,barikeri-etal-2021-redditbias,jentzsch-turan-2022-gender}.
However, these methods typically require manually created debiased data, which is resource-intensive.

In this work, we aim to mitigate biases and stereotypes of LMs
(hereafter referred to collectively as ``bias'')
using a proposed method inspired by the task arithmetic approach~\cite{ilharco2023editing}.
We hypothesize that biases can be reduced
through vector subtractions in the parameter space,
assuming the same model architecture for all LMs.
Most existing debiasing techniques rely on manually created debiased data.
In contrast,
our proposed debiasing method avoids the necessity of resource-intensive manual work.
Specifically, we construct debiased LMs by subtracting the ``Bias Vectors'' from the weights of LMs.

An overview of the proposed method is shown in Figure~\ref{fig:overview}.
The Bias Vector is created by subtracting
the weights of a pre-trained LM
from 
those of a biased LM,
which is continually trained on biased text.
By applying this Bias Vector to the pre-trained LM weights,
we construct a debiased LM.

The masked language modeling (MLM) task is adopted
for the continual training,
and our experiments target BERT~\citep{devlin-etal-2019-bert},
ALBERT~\citep{Lan2020ALBERT},
and RoBERTa~\citep{liu2019roberta},
following \citet{meade-etal-2022-empirical}.
We evaluated the debiased LMs using the Sentence Encoder Association Test (SEAT)
\citep{may-etal-2019-measuring}
and confirmed the effectiveness of our Bias Vector method.

Additionally, we analyzed how scaling the Bias Vector by a factor $\lambda$ influences LM biases,
allowing us to control the magnitude of the Bias Vector applied to LMs.  
Furthermore, the evaluation on the GLUE benchmark \citep{wang-etal-2018-glue}
demonstrated that LM representations remain effective for downstream tasks,
even after applying the Bias Vector with $\lambda=1$.

Our main contributions are as follows:
\begin{itemize}
  \item Proposing the ``Bias Vector'' method, that enables debias LMs development without manually creating debiased data;
  \item Verifying the effectiveness of the Bias Vector method
  and confirming that
  debiased LMs have equivalent performance as pre-trained LMs on GLUE;
  \item Confirming that over-debiasing (i.e., with large $\lambda$) can lead to a collapse of pre-trained knowledge,
  by analyzing the effect sizes on SEAT and the GLUE scores.

\end{itemize}

\section{Related Works}
\subsection{Language Models and Bias}
Language models (LMs) are inherently biased
because their training processes rely on human-created text data,
which would reflect human biases
\citep{NIPS2016_a486cd07}.
\citet{Navigli2023} defined
the term \textit{bias} in the field of Natural Language Processing as
``prejudices, stereotypes, and discriminatory attitudes against certain groups of people.''
We adopt this bias definition throughout this paper.

Various debiasing methods
have been proposed to mitigate these biases
\citep{schick-etal-2021-self,zmigrod-etal-2019-counterfactual,Webster2020,ravfogel-etal-2020-INLP,liang-etal-2020-towards}.

Several studies have shown that for word-embedding models, such as word2vec \citep{mikolov2013w2v},
the bias in word embeddings can be mitigated using approaches like
subtracting the statistically significant mean vector associated with the bias from each word vector
\citep{NIPS2016_a486cd07,mu2018allbutthetop,gonen-goldberg-2019-lipstick,wang-etal-2020-double}.
In contrast, other studies ahve proposed bias mitigation techniques specifically
for Transformer-based LMs
\citep{ravfogel-etal-2020-INLP,liang-etal-2020-towards}.

Several benchmarks have been introduced to evaluate debiasing approaches.
\citet{weat2016} developed the
Word Embedding Association Test (WEAT) to measure
bias scores in word embeddings.
\citet{may-etal-2019-measuring} proposed the Sentence Encoder Association Test (SEAT)
as an extension of WEAT, extending the focus from word to sentence.
StereoSet \citep{nadeem-etal-2021-stereoset} is another benchmark designed
to evaluate stereotypes across four bias categories:
\textit{race}, \textit{profession}, \textit{gender}, and \textit{religion}.
StereoSet consists of two subsets:
\textit{intrasentence}, which measures biases within a individual sentence,
and \textit{intersentence}, which evaluates biases at the discourse level across multiple sentences.
\citet{nangia-etal-2020-crows} also introduced the CrowS-Pairs benchmark for bias neasurements.

However, \citet{meade-etal-2022-empirical} criticized existing debiasing methods,
arguing that these methods have focused too narrowly on their effectiveness within specific datasets.
Therefore, they conducted an experimental evaluation of
these methods using specific LMs 
on bias benchmarks
and released the evaluation code for debiasing approaches.
We utilized the code\footnote{\url{https://github.com/mcgill-nlp/bias-bench}.}
in our evaluation experiments.

\subsection{Task Arithmetic Approaches}
Recent studies have focused on the weight manipulation weights in neural network models
\citet{ilharco2023editing}.
Several approaches for merging model weights
have been proposed in the field of Computer Vision, 
\citep{wortsman2022modelsoups,michael2022Merging,ainsworth2023gitrebasinmergingmodels}.
\citet{wortsman2022modelsoups} found that
a model constructed by averaging the weights of multiple models
fine-tuned with different hyperparameters
often results in improved model performance and robustness.
\citet{michael2022Merging}
proposed that
computing the average parameter weights in different models
corresponds to approximating
the posterior distribution of each model parameter
\citet{michael2022Merging} proposed a method to combine the characteristics of
each model by considering the mean of multiple model parameters with the same architecture.
\citet{ainsworth2023gitrebasinmergingmodels}
hypothesized that the loss landscape in the training and optimization process
of deep learning models
exhibits a ``single basin'' phenomenon
and introduced an algorithm to align the weights between models.

Some studies in Natural Language Processing
have also attempted to manipulate LM weights.
\citet{li2022branchtrainmergeembarrassinglyparalleltraining}
improved the overall LM performance by dynamically updating
and merging multiple expert LMs that were independently trained on different data subsets;
therefore, the LMs could be effectively trained toward domain-specific knowledge.

Inspired by these works,
\citet{ilharco2023editing} introduced the task arithmetic approach,
which edites model parameters
using a task vector containing the information necessary to achieve good performance on a given task.
Motivated by the task arithmetic concept,
\citet{huang-etal-2024-chat} introduced the ``Chat Vector'' approach
which enables pre-trained LMs
to gain conversational abilities in new languages
without any additional training.

\citet{ilharco2023editing} also evaluated the toxicity of LMs;
however, their evaluation results
did not align with the benchmarks for bias evaluation.
In addition, its effectiveness was demonstrated with only GPT-2 model \citep{radford2019language}.
We comprehensively examine the effectiveness
of our Bias Vector method in the bias benchmarks following \citet{meade-etal-2022-empirical}.

\section{Proposed Methods}
\subsection{Continual Training} \label{chap3-1}
We continually train the LMs using biased text data to adjust their parameters toward the biased LMs.

As an additional training task,
we adopt the masked language modeling
(MLM)\footnote{Our MLM experiments follow the HuggingFace library:
\url{https://huggingface.co/docs/transformers/main/en/tasks/masked_language_modeling}.},
which is also used in the BERT pre-training process.

In MLM task,
a portion of tokens in sentences is replaced with [MASK] tokens,
and LMs are trained to predict these masked tokens.

\subsection{Bias Vector} \label{chap3-2}
In order to mitigate biases in LMs,
we propose the ``Bias Vector'' method, inspired by the task arithmetic approach \citep{ilharco2023editing},
assuming the LMs share the same model architecture.
An overview of the proposed method is presented in Figure~\ref{fig:overview}.

Our experiments are conducted using pre-trained LMs such as BERT \citep{devlin-etal-2019-bert}.
We continually train these LMs on biased text data, following the process described in Section~\ref{chap3-1}.

We construct a Bias Vector
by subtracting the weights of the biased LMs
from
those of the pre-trained LMs.
This process can be described by the following equation:

\begin{equation}
  V_{bias} = \theta_{bias} - \theta_{org},
\end{equation}
where $\theta_{org} \in \mathbb{R}^{\mathrm{d}}$
and $\theta_{bias} \in \mathbb{R}^{\mathrm{d}}$
represent the weights of pre-trained LMs and biased LMs, respectively,
and
$V_{bias} \in \mathbb{R}^{\mathrm{d}}$ is the Bias Vector.
Here, the LM parameters and the Bias Vector are represented in $d$ dimensions.
Since the pre-trained LMs and biased LMs are composed of the same model structures,
their parameters $\theta_{org}$, $\theta_{bias}$, and $V_{bias}$
can be directly added to or subtracted from one another.

In addition to calculating the Bias Vector,
we construct debiased LMs
by subtracting the Bias Vector from the pre-trained LM weights. 
This procedure is represented by the following equation:

\begin{equation}
  \label{eq:debias}
  \theta_{debias} = \theta_{org} - \lambda ~V_{bias},
\end{equation}
where $\theta_{debias} \in \mathbb{R}^{\mathrm{d}}$ denotes the weights of the debiased LMs,
which share the same architecture as $\theta_{org}$,
The hyperparameter $\lambda \in R$ is a scaling factor used to control the magnitude of the Bias Vector.

In this subtraction process,
the Layer Normalization layers are excluded from the parameters to be subtracted.
These layers are designed to solely normalize the data distribution and do not learn any bias information.

\section{Experiments}

\subsection{Target Pre-trained LMs}
In our experiments, we adopt three LMs:
BERT \citep{devlin-etal-2019-bert},
ALBERT \citep{Lan2020ALBERT},
and RoBERTa \citep{liu2019roberta}.
These LMs are chosen based on the empirical survey by
\citet{meade-etal-2022-empirical} for bias investigation.

The links to these pre-trained LMs are listed in Appendix~\ref{sec:lm-url}.

\begin{table}[t]
  \centering
  \begin{tabular}{|c|p{5cm}|}
    \hline
    bias & \multicolumn{1}{c|}{text}\\ \hline \hline
    race &
      The mountain tribes of \textcolor{red}{afghanistan} have a reputation for being the most \textbf{dangerous} peoples on earth.\\ \hline
    gender &
      The \textcolor{red}{mother} takes care of the \textbf{children} at home.\\ \hline
    profession &
      The \textcolor{red}{civil servant} was a \textbf{bureaucrat} at heart, so he knew where he really belonged.\\ \hline
    religion &
      The \textcolor{red}{bible} is \textbf{holy} scripture.\\ \hline
  \end{tabular}
  \caption{Examples of the StereoSet intrasentence dataset used for the continual training.
  This dataset consists of sentences with one word blanked out and an associated bias type
  (\textit{race}, \textit{profession}, \textit{gender}, and \textit{religion}).
  In these examples, 
  \textbf{bold words} in text indicate blanked-out words in the original StereoSet dataset,
  and \textcolor{red}{red words} represent the targets of stereotypes.}
  \label{tab:stereoset}
\end{table}

\begin{table}[th]
  \centering
  \begin{tabular}{|c|c|p{3.5cm}|}
    \hline
    type & category & \multicolumn{1}{c|}{text}\\ \hline \hline
    target & Science & The experiment is here.\\ \cline{3-3}
     & & The person's name is Einstein.\\ \cline{2-3}
     & Arts & This is a symphony.\\ \cline{3-3}
     & & The dramas are here.\\ \hline
    attribute & Female & That is a mother.\\ \cline{3-3}
     & & This is a grandmother.\\ \cline{2-3}
     & Male & That is a father.\\ \cline{3-3}
     & & This is a grandfather.\\ \cline{2-3}
    \hline
  \end{tabular}
  \caption{
    Examples of SEAT dataset for evaluating social bias.
    These samples are a subset of SEAT-8 data used to evaluate the \textit{gender} bias.}
  \label{tab:seat_ex}
\end{table}

\subsection{Experimental Setup for Continual Training}

In this section, we outline the details of the continual training for building biased LMs.

\subsubsection{Training Dataset}
We utilize the StereoSet intrasentence dataset \citep{nadeem-etal-2021-stereoset}
for the continual training
of the target LMs in our experiments.
The dataset consists of biased text categorized into four types
(\textit{race}, \textit{profession}, \textit{gender}, and \textit{religion}),
sentences with one word blanked out,
and a set of options for a \textit{fill-in-the-blank} task.
These options include three types of words: stereotype, anti-stereotype, and meaningless.

To construct a bias-only dataset for the continual training
we fill the blanks with stereotype options (i.e., a biased word).
The other options are excluded from the continual training process.
Examples from this dataset are presented in Table~\ref{tab:stereoset}.

This dataset for the continual training contains 8,498 sentences, categorized as follows:
\textit{race} (3,989), \textit{profession} (3,269),
\textit{gender} (996), and \textit{religion} (604).
In our experiments, 15\% of the tokens in the text
are randomly masked with [MASK] tokens for the MLM task.

It is important to note that the StereoSet intrasentence dataset
reflects stereotypes as perceived by annotators who were residents of the United States (U.S.).
Since stereotypes vary not only by gender and race
but also by cultural and regional contexts
\citep{nadeem-etal-2021-stereoset},
the biases that can be mitigated using our proposed method are limited to
those biases held by the U.S. annotators.

\subsubsection{Experimental Details}
Our experiments are conducted
with the following hyperparameters.
We use
AdamW~\citep{loshchilov2018decoupled} as the optimizer, which improves weight decay behavior over Adam~\citep{kingma2017adammethodstochasticoptimization}.
The learning rate is set to 1e-4,
the weight decay is 0.01,
the number of warmup steps is fixed to 10,000,
the batch size is 128,
and the learning rate scheduler is linear.
All other training parameters
follow the default settings provided by
the Training Arguments library.\footnote{\url{https://huggingface.co/docs/transformers/v4.40.2/en/index}.}
To effectively overfit the LMs toward biases,
we train the models with the number of epochs set to 30.

We construct the Bias Vectors using ten different seeds,
and evaluate the average effect sizes of our debiasing method.
The seed values remain consistent across all evaluation experiments.

The scaling factor $\lambda$ of the Bias Vector is set to 1, 10, or 100
to analyze how varying magnitudes of the vector impact bias mitigation.

The computational resources used for the continual training
are described in Appendix~\ref{sec:computing}.

\subsection{Experimental Setup for Debias Evaluation}\label{sec:experiments_debias_eval}
This section describes the experimental setup for
evaluating the debiasing methods.

\subsubsection{Debias Benchmark}
Our experiments used the Sentence Encoder Association Test (SEAT) \citep{may-etal-2019-measuring}
to evaluate the bias magnitudes of the debiased LMs,
following \citet{meade-etal-2022-empirical}.

The SEAT is an extension of the Word Embedding Association Test (WEAT) \citep{weat2016}
to measure LM biases in sentence embeddings.
WEAT comprises two sets of \textit{attribute} words and two sets of \textit{target} words.
For example, the \{female / male\} \textit{attribute} sets
and the \{science / arts\} \textit{target} sets
can be used to evaluate biases, such as \textit{gender}-related bias.

Table~\ref{tab:seat_ex} shows examples from SEAT-8,
a subset specifically designed to evaluate \textit{gender} bias as part of social biases.

It should be noted that
the StereoSet dataset,
used for the continual training (Section~\ref{chap3-1}),
is excluded from our evaluation experiments to prevent data leakage.
Instead, we rely on the SEAT benchmark, which provides an assessment of bias magnitudes.

\subsubsection{Evaluation Metrics}
This section explains the bias evaluation metrics for assessing LMs.

The bias magnitude is measured based on
the statistical method Cohen's d
which calculates the effect sizes of two groups as follows:

\begin{align}\label{eq:d}
    d = \frac{\mathrm{diff}(X,Y,A,B)}{\sigma\left( \{s(t, X, Y) \mid t \in A \cup B\} \right)}.
\end{align}
where, $\mu$
represents the mean,
and $\sigma$ denotes the standard deviation.
$A$ and $B$ are sets of \textit{attribute} sentences,
and $X$ and $Y$ are sets of \textit{target} sentences.
$\mathrm{diff}(X,Y,A,B)$
is the result of subtracting
$\mu\left(\{s\left(y,A,B\right)\mid y\in Y\}\right)$
from $\mu\left(\{s\left(x,A,B\right)\mid x\in X\}\right)$.

Here, $s(w,A,B)$ is the difference in cosine similarities
between a sentence $w$ and the sets $A$ and $B$:

\begin{align}
    &s(w,A,B) \nonumber \\
    &= \frac{1}{|A|} \sum_{a \in A} \cos(w, a) - \frac{1}{|B|} \sum_{b \in B} \cos(w, b).\label{eq:s-wab}
\end{align}

We evaluate our debiasing approach on SEAT by Equation~\ref{eq:d}.

\subsection{Experimental Setup for GLUE}\label{sec4-glue}
To ensure that our debiasing method does not degrade the effectiveness of LM representations,  
we evaluate both our debiased LMs and the pre-trained LMs
on the GLUE benchmark \citep{wang-etal-2018-glue} after fine-tuning.
The training data is randomly split into a 9:1 ratio:
90\% is used for training and 10\% for validation.
The original validation data from the GLUE benchmark are used as test data.

The computational resources are provided in Appendix~\ref{sec:computing}.
We determine the hyperparameters for fine-tuning as described
in Appendix~\ref{app:glue}.

\subsection{Baselines}
We adopt the pre-trained LMs (BERT, ALBERT, and RoBERTa) as baselines
to measure the effect of our debiasing methods.  
These baselines provide a point of comparison to assess both
bias mitigation and the preservation of downstream task performance.

\begin{table}[t]
  \centering
  \scalebox{0.87}{
  \begin{tabular}{l|rrr}
    \hline
    \multicolumn{1}{c|}{Methods} &  \multicolumn{1}{c}{BERT} & \multicolumn{1}{c}{ALBERT} & \multicolumn{1}{c}{RoBERTa}
    \\ \hline \hline
    Pre-Trained LM & 0.672 & 0.675 & 0.733 \\
    ~w/ BV(race, $1$) & 0.646 & 0.663 & 0.657 \\
    ~w/ BV(prof., $1$) & 0.661 & 0.683 & 0.657 \\
    ~w/ BV(gender, $1$) & 0.653 & 0.736 & 0.672 \\
    ~w/ BV(religion, $1$) & 0.652 & 0.735 & 0.671 \\
    ~w/ BV(all, $1$) & 0.447 & 0.534 & 0.570 \\
    ~w/ BV(all, $10$) & 0.446 & 0.311 & \textbf{0.272} \\
    ~w/ BV(all, $100$) & \textbf{0.202} & \textbf{0.201} & 0.411 \\ \hline
  \end{tabular}
  }
  \caption{
    Average scores of absolute effect sizes across
    \textit{gender}, \textit{race}, and \textit{religion}
    using pre-trained or debiased LMs (BERT, ALBERT and RoBERTa).
    BV(\textit{bias}, $\lambda$) refers to the Bias Vector utilizing
    \textit{bias}-typed data with the scaling factor set to $\lambda$.
    The abbreviation ``prof.'' stands for \textit{profession}, indicating a specific bias type.
    Effect sizes closer to 0 suggest that LM representations are less biased.
    }
  \label{tab:results}
\end{table}

\begin{figure}[t]
  \includegraphics[width=\columnwidth]{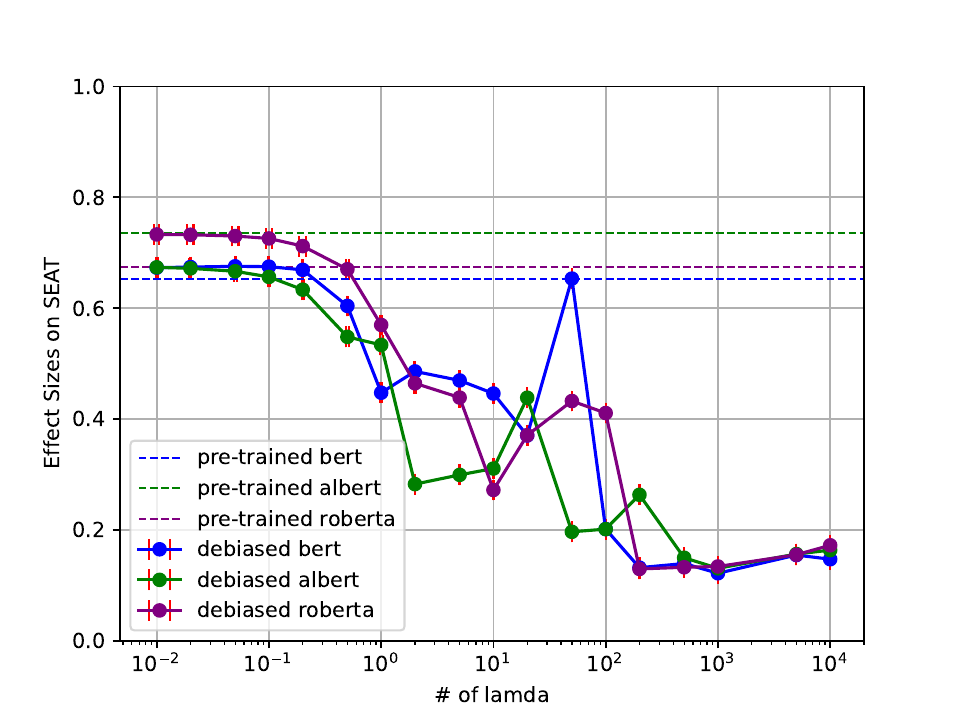}
  \caption{
  Variation of effect sizes on the SEAT with the scale factor $\lambda$.
  The dashed lines indicate the effect sizes on pre-trained LMs.
  The closer the effect size is to zero, the smaller the bias.}
  \label{fig:lambda}
\end{figure}

\begin{table*}[th]
  \centering
  \scalebox{0.92}{
    \begin{tabular}{l|ccccccccccc}
      \hline
      Methods
      & cola & sst-2 & mrpc & sts-b & qqp & mnli-\{m/mm\} & qnli & rte & wnli & avg. \\
      \hline\hline
      BERT
      & 0.556 & 0.927 & 0.830 & 0.854 & 0.891 & 0.837/0.835 & 0.906 & 0.523 & 0.366 & 0.776 \\
      ~w/ BV(all, 1)
      & 0.572 & 0.922 & 0.821 & 0.851 & 0.892 & 0.840/0.838 & 0.896 & 0.585 & 0.352 & \textbf{0.779} \\
      ~w/ BV(all, 10)
      & 0.000 & 0.894 & 0.763 & 0.858 & 0.845 & 0.823/0.828 & 0.860 & 0.505 & 0.408 & 0.678 \\
      ~w/ BV(all, 100)
      & 0.000 & 0.509 & 0.748 & 0.038 & 0.316 & 0.327/0.330 & 0.495 & 0.473 & 0.437 & 0.367 \\
      \hline
  
      ALBERT
      & 0.508 & 0.923 & 0.822 & 0.870 & 0.888 & 0.842/0.849 & 0.913 & 0.520 & 0.408 & 0.779
      \\
      ~w/ BV(all, 1)
      & 0.545 & 0.920 & 0.770 & 0.866 & 0.888 & 0.844/0.847 & 0.907 & 0.570 & 0.521 & \textbf{0.785}\\
      ~w/ BV(all, 10)
      & 0.000 & 0.849 & 0.748 & 0.697 & 0.316 & 0.327/0.330 & 0.495 & 0.527 & 0.437 & 0.473 \\
      ~w/ BV(all, 100)
      & 0.000 & 0.509 & 0.748 & 0.090 & 0.316 & 0.327/0.330 & 0.495 & 0.473 & 0.437 & 0.373 \\
      \hline
  
      RoBERTa
      & 0.552 & 0.944 & 0.763 & 0.891 & 0.877 & 0.879/0.873 & 0.924 & 0.527 & 0.563 & \textbf{0.794}\\
      ~w/ BV(all, 1)
      & 0.539 & 0.944 & 0.758 & 0.869 & 0.899 & 0.875/0.873 & 0.928 & 0.527 & 0.563 & 0.792\\
      ~w/ BV(all, 10)
      & 0.000 & 0.919 & 0.850 & 0.884 & 0.890 & 0.869/0.867 & 0.910 & 0.625 & 0.563 & 0.737 \\
      ~w/ BV(all, 100)
      & 0.000 & 0.509 & 0.748 & 0.015 & 0.316 & 0.327/0.330 & 0.495 & 0.527 & 0.437 & 0.370 \\
      \hline
      \end{tabular}
  }
  \caption{\label{tab-glue}
    GLUE evaluation scores with fine-tuning pre-trained LMs and the debiased LMs with Bias Vector methods ($\lambda=1$).
    To save space in this table,
    the results of MNLI-matched and MNLI-mismatched are displayed in the same cell (matched / mismatched),
    and cells of \{MRPC, QQP\} show average scores over accuracies and F1 scores.
    STS-b cells show average values of pearson and spearman correlations.
    Again, BV(\textit{bias}, $\lambda$) refers to the Bias Vector utilizing
    \textit{bias}-typed data with the scaling factor set to $\lambda$.
  }
\end{table*}

\section{Results and Discussion}
We evaluated the LMs using Bias Vectors constructed with the following bias type data:
\textit{race}, \textit{profession}, \textit{gender}, \textit{religion},
and a combination of all these types (\textit{all}).

\subsection{SEAT Results} \label{sec:5-1}
The results of effect sizes on SEAT are shown in Table~\ref{tab:results}.
BV(\textit{bias}, $\lambda$) refers to the Bias Vector utilizing
\textit{bias}-typed data with the scaling factor set to $\lambda$.

The effect sizes of the debiased BERT with $\lambda=1$
improved by 0.224 points
compared to the baseline model (average over ten seed values).

Performance improvements were consistently observed for the other LMs.
Compared to the baselines,
ALBERT improved by 0.142 points,
and RoBERTa achieved a gain of 0.164 points (with $\lambda=1$).

On average,
the proposed method achieved a 0.177 point improvement across the three LMs,
demonstrating consistent bias mitigation effectiveness.

Specifically, we investigated
the relationship between effect sizes
and $\lambda$.
Table~\ref{tab:results} and Figure~\ref{fig:lambda}
show that increasing $\lambda$ reduces the effect size on SEAT,
indicating that larger scaling factors lead to further bias mitigation.

\subsection{GLUE Scores}\label{sec:5-2}
We investigated the impact of the Bias Vector
on the performance of downstream tasks in the GLUE benchmark
by comparing the debiased and pre-trained LMs.
Table~\ref{tab-glue} reports the GLUE scores for fine-tuned pre-trained LMs
and debiased LMs using the Bias Vector method with $\lambda=1$.
Compared with the pre-trained LMs,
the debiased models showed average performance improvements of 0.23\%
(BERT: 0.3\%, ALBERT: 0.6\%, and RoBERTa: -0.2\%).

Our approach does not harm LM representations when $\lambda=1$,
allowing the LMs to maintain high performance after fine-tuning even after debiasing.
The slight reduction observed in RoBERTa probably falls within the margin of error,
suggesting a minimal impact on its performance.

With a larger scaling factor $\lambda=10$,
performance degradation was minimal for BERT and RoBERTa in most tasks.
ALBERT exhibited a notable decline, averaging 30.6\%.

For the CoLA dataset,
the performances notably decreased,
indicating the need for further investigation into task-specific effects.

When the scaling factor was excessively increased ($\lambda=100$),
the GLUE scores significantly declined.
These results suggest that overly increasing the scaling factor
severely damages the representational capabilities acquired during pre-training.

\subsection{SEAT Results on \textit{Profession} Bias}\label{sec:5-3}
The SEAT data do not strictly
evaluate the \textit{profession} bias \citep{meade-etal-2022-empirical}.
Since the bias data used for the continual training in this study includes the \textit{profession} bias,
this section investigates the effects of incorporating this bias into the training process.

The bias mitigation was also observed with BV(\textit{prof.}, 1)
in the SEAT results as shown in Table~\ref{tab:results}.
The effect sizes for RoBERTa improved from 0.733 (pre-trained)
to 0.657 (debiased with BV(\textit{prof.}, 1)).

Since different types of biases are interrelated,
debiasing \textit{profession} bias likely mitigates other biases as well.
For instance,
the sentence ``Engineers are male''
reflects both \textit{profession} and \textit{gender} biases.
If such \textit{profession}-biased sentences are learned
during the construction of BV(\textit{prof.}, 1),
the resulting Bias Vector may unintentionally encoded other biases,
contributing to the improved effect sizes.

The occurrence of bias duplication highlights
the need for task arithmetic approaches
that prevent overlapping bias vectors from being subtracted multiple times.
For example, removing both the \textit{profession} and \textit{gender} Bias Vectors
from a pre-trained LM may inadvertently amplify mitigation effects, leading to over-debiasing.

Future work should focus on developing methods to address overlapping biases more effectively,
ensuring precise bias mitigation across multiple biases.

\subsection{\texorpdfstring{Effectiveness of $\lambda$}{}}\label{sec:5-4}
To evaluate the effectiveness of the scaling factor $\lambda$,
we varied its value from 0.01 to 10,000
and measured the resulting effect sizes on SEAT.
The results are reported in Figure~\ref{fig:lambda}.

The evaluation across all SEAT datasets confirmed that
the effect sizes converged approximately to zero.

Our initial hypothesis was that
increasing the scale factor $\lambda$ of the Bias Vector
would first reduce the effect size (debiasing),
and then shift it toward an anti-stereotypical effect size (biasing).

For instance, if the Bias Vector had been learned in the \textit{male} direction,
we expected that increasing the scale factor $\lambda$
would gradually be biased in the \textit{female} direction.

Contrary to this hypothesis, the results showed that the effect size consistently converged toward zero across all evaluations.
This outcome may be due to two possible reasons:
(1) The task arithmetic approach used to construct the Bias Vector
may not have effectively captured the specific bias direction (investigating in Section~\ref{sec:5-5}); and
(2) The biased LM may learn unintended information during continual training,
leading to a collapse in the representations of the debiased LM when $\lambda$ is scaled up (discussing in Section~\ref{sec:5-6}).

\begin{figure}[t]
  \centering
  \includegraphics[width=8cm]{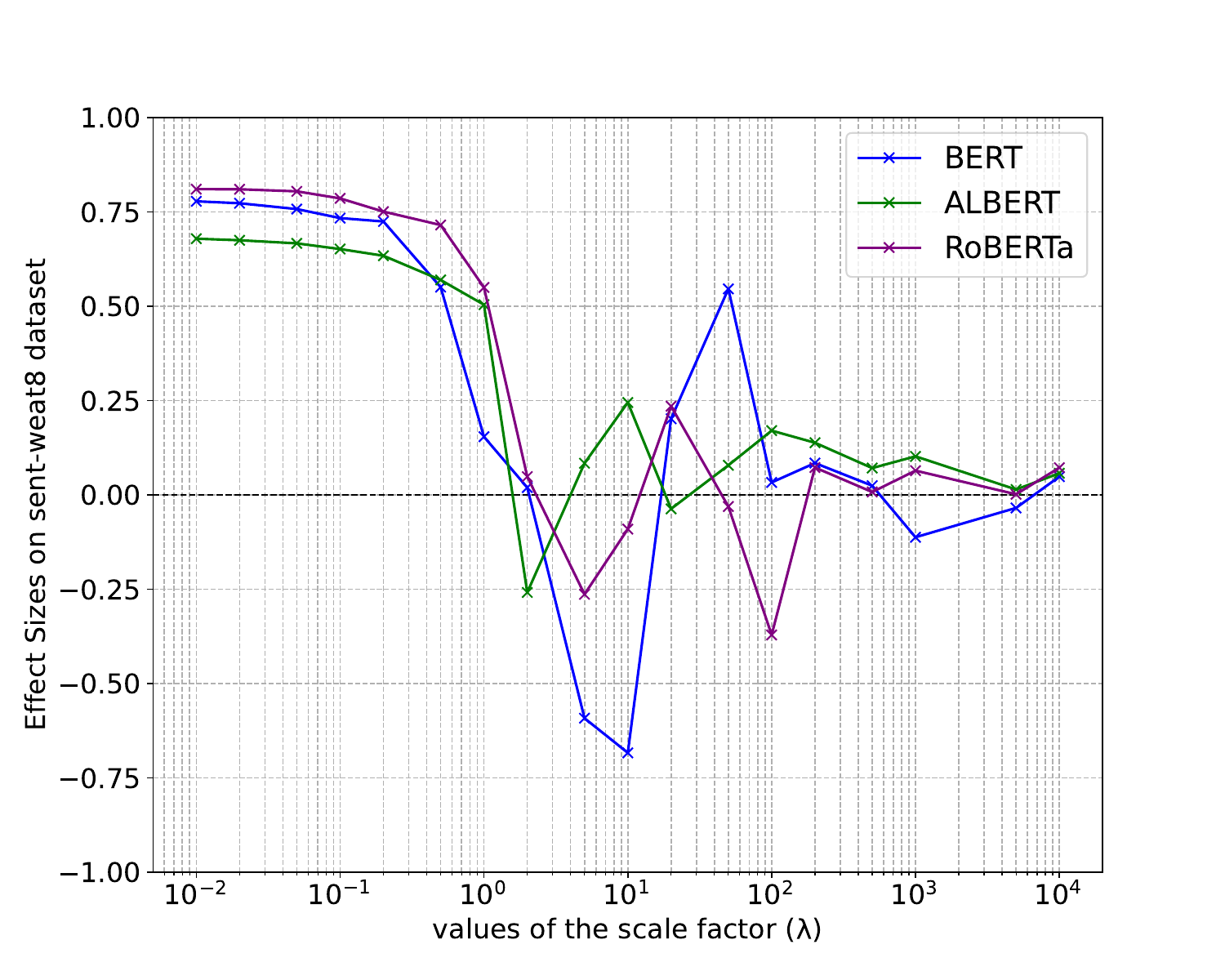}
  \caption{Effect sizes on the \textit{gender}-biased SEAT dataset (SEAT-8) with varying $\lambda$.
  The effect sizes are computed as the average of scores across ten different seed values.
  The closer the effect size is to zero, the smaller the bias.}
  \label{fig:all_lm_seat8}
\end{figure}

\begin{figure}[t]
  \centering
  \includegraphics[width=8cm]{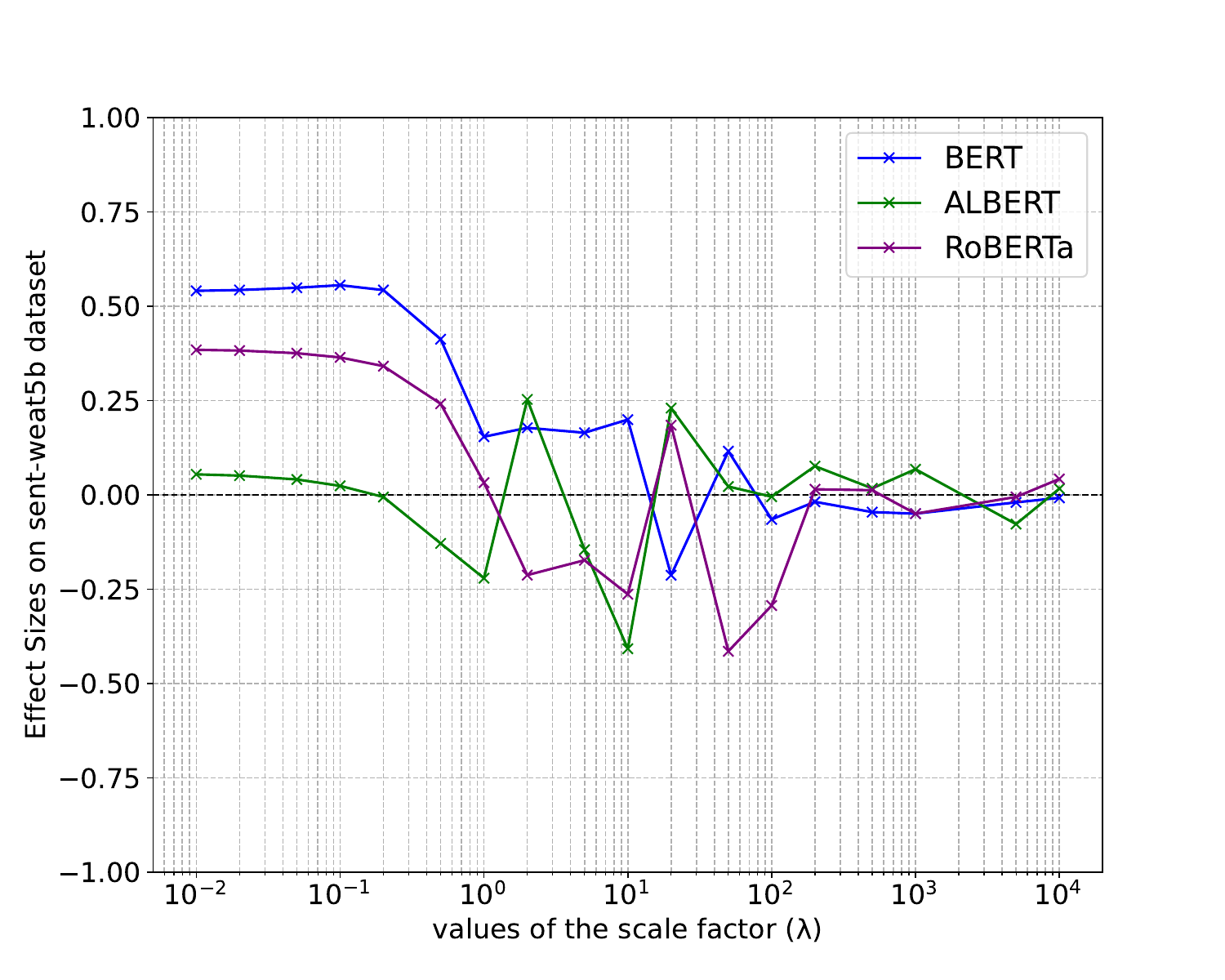}
  \caption{Effect sizes on the \textit{race}-biased SEAT dataset (SEAT-5b) with varying $\lambda$.
  The effect sizes are computed as the average of scores across ten different seed values.
  The closer the effect size is to zero, the smaller the bias.}
  \label{fig:all_lm_seat5b}
\end{figure}

\subsection{Effect Size Behavior in Each SEAT Task}\label{sec:5-5}
In Section~\ref{sec:5-4}, we hypothesized that effect sizes would initially decrease (debiasing)
and then increase in the opposite direction.
However, the observed results deviated from this trend.

This section investigates whether the discrepancy between our hypothesis and the SEAT results
can be attributed to the Bias Vectors failing to adequately capture bias information.
To explore this issue, we analyze the effect sizes for each bias category within the SEAT dataset.

The effect sizes for SEAT-8 (\textit{gender} bias) and SEAT-5b (\textit{race} bias)
are shown in Figure~\ref{fig:all_lm_seat8} and Figure~\ref{fig:all_lm_seat5b}, respectively.
The corresponding results, including standard deviations over ten different seed values, are shown in Appendix~\ref{app:seat_dev}.

For ALBERT,
the effect sizes begin to converge toward zero after $\lambda$ exceeds one,
whereas for other LMs,
the scores approach zero after $\lambda$ is larger than 10.

For each model, prior to convergence,
we observed behavior consistent with our hypothesis:
initial debiasing occurs, followed by a reversal of the bias direction
(i.e., effect sizes increase in the anti-stereotypical direction).
The increase in effect sizes in the opposite direction confirms that
the Bias Vector successfully mitigates the biases in LMs,
demonstrating that its training process effectively captured the intended bias direction.

These findings confirm that the Bias Vector effectively mitigates
biases and captures the intended bias direction.

\subsection{\texorpdfstring{Impact of $\lambda$ on LM Representations}{}}\label{sec:5-6}
According to Section~\ref{sec:5-4},
the effect sizes approached zero as $\lambda$ increased.

The convergence behavior of the effect sizes varies across LMs.
For BERT and RoBERTa, the convergences occur when $\lambda$ is set between 10 and 100,
while, the convergence begins around $\lambda=10$ for ALBERT.

As mentioned in Sections~\ref{sec:5-2} and \ref{sec:5-5},
the collapse of representations in LMs
was observed at $\lambda=10$ for ALBERT
and at $\lambda=100$ for BERT and RoBERTa.

This observation suggests that the convergence of the effect sizes toward zero
coincides with a collapse in the LM representations across all models.
Specifically, as $\lambda$ increases,
the LMs lose their ability to distinguish between stereotypical and anti-stereotypical information,
leading to predictions that are uniformly inaccurate.
This inaccuracy reduced the difference in effect sizes between the two types of information,
leading to a false impression of sufficient bias mitigation.

These results indicate that
the small effect sizes observed for large values of $\lambda$ do not signify successful bias mitigation.
Rather, they reveal a collapse in the LM representations at large $\lambda$,
where the models fail to distinguish between stereotypical and anti-stereotypical information.
Consequently, LM predictions become inaccurate for both types of information, driving the bias effect sizes toward zero.


\section{Conclusions and Future Works}
In this paper, we introduced a ``Bias Vector'' method
for bias mitigation of language models (LMs)
without manually created debiasing data.
We constructed the Bias Vector
by calculating the difference between the weights of the pre-trained LMs
and
those of the biased LMs,
which were continually trained on the biased text.
We attempted to mitigate the LM bias by
subtracting the Bias Vector from the pre-trained LM weights.

On average over three LMs (BERT, ALBERT, and RoBERTa),
our debiasing method improved 0.177 points on all test sets in SEAT
with setting the scale factor $\lambda=1$.
We also confirmed that the debiased LMs
using our method had an average score improvement of 0.23\% on the GLUE benchmark.
These results demonstrate that our method can successfully debias LMs
with preserving their representational performances.

By varying $\lambda$ from 0.01 to 10,000, we observed that effect sizes decreased and approached zero.
However, for large $\lambda$ values (e.g., $\lambda=100$), the GLUE scores significantly declined,
suggesting that this bias mitigation may result
from a collapse of pre-trained knowledge rather than the effectiveness of our method.

Future work will focus on further analyzing the relationship between
the scaling factor $\lambda$ and SEAT scores to better understand the behavior of bias mitigation.
Additionally, given the widespread use of Large Language Models (LLMs),  
we aim to extend the Bias Vector approach to LLMs and evaluate
its effectiveness on these models.

\section*{Limitations}
In this study, 
we evaluated debiased LMs on GLUE benchmark
to ensure that LM representations
had not decreased compared to pre-trained LMs
by our debias methods ``Bias Vector.''
This paper presented only
the GLUE scores using our debiased LMs with $\lambda=1, 10, 100$.
Evaluations of debiased LMs on other $\lambda$ conditions are not conducted
due to limited computational resources.
To confirm the relationship between $\lambda$ and GLUE scores,
the GLUE evaluation experiments on the other $\lambda$
should be conducted in future.

Following \citet{meade-etal-2022-empirical},
we should evaluate our method toward GPT-2 model,
in addition to BERT, ALBERT and RoBERTa.
However, due to computational resource constraints,
GPT-2 was not conducted in our experiments.
We plan to conduct and evaluate those experiments in the future.

\section*{Ethics Statement}
\citet{Navigli2023} defined
the term \textit{bias} in the field of Natural Language Processing as
``prejudices, stereotypes, and discriminatory attitudes against certain groups of people.''
We adopt this bias definition throughout this paper.

For this bias definition,
we refer to both stereotypes and biases as ``bias'' for simplicity.
We understand that these are different concepts,
and we acknowledge that the stereotypical data (StereoSet) 
used in our experiments reflect those of the U.S. residents
\citep{nadeem-etal-2021-stereoset}.

We particularly address bias mitigation for LMs
by utilizing stereotypes.
Biases
arise when concepts that should not be associated with
particular social groups are unfairly linked (e.g., ``programmers are male'').
If LLM systems possess such biases,
they are likely to leave a negative impression on users.
This work examines the applicability of a task arithmetic approach
for bias mitigation.
The purpose of our study is to reduce the LM bias using the proposed methods.

We understand the importance of maintaining
an objective stance.
Therefore,
we emphasize that the content of this study
is not influenced by our political positions, stereotypes or biases.
Our research aims to respect
the ethical principle of fairness in scientific inquiry
and make responsible and constructive contributions
to the development of AI technologies.

\section*{Acknowledgments}

Our profound appreciation is express our profound gratitude
to the anonymous reviewers for their thorough comments,
and to Prof. Yasutomo Kimura in Otaru University of Commerce for
his expertise and insightful feedback.

We would like to extend our sincere appreciation to Mr. Koji Tanaka and Mr. Tatsuhiko Saito
in Mitsubishi Electric Corporation
for their unwavering support throughout this research endeavor.

\bibliography{custom}

\appendix

\section{Language Models}\label{sec:lm-url}
For evaluating our methods, we adopt three LMs:
BERT \citep{devlin-etal-2019-bert}
ALBERT \citep{Lan2020ALBERT},
and RoBERTa \citep{liu2019roberta}.
These models are available on the following sites:

\begin{itemize}
  \item BERT: \url{https://huggingface.co/google-bert/bert-base-uncased};
  \item ALBERT: \url{https://huggingface.co/albert/albert-base-v2};
  \item RoBERTa: \url{https://huggingface.co/FacebookAI/roberta-base}.
\end{itemize}

\section{Computing Environments}\label{sec:computing}
The process of generating biased LMs and our proposed Bias Vector
was facilitated using four GPUs (NVIDIA RTX A6000),
a procedure that spanned several hours.
In the same way,
the GLUE training procedure,
which was conducted without the exploration of hyperparameter combinations,
required approximately a full day
utilizing four GPUs (NVIDIA Quadro RTX 8000).

\section{Experimental Setup for GLUE} \label{app:glue}

\subsection{Training Arguments for BERT} \label{app:glue_bert}
In addition to ALBERT, we fine-tune BERT for GLUE downstream tasks.
We determine hyperparameters following \citet{devlin-etal-2019-bert},
i.e., we explore all combinations of the following hyperparameters and evaluate the model,
which yields the best score on the validation dataset, using the test data on each task.

\begin{itemize}
  \item Batch size: 16, 32
  \item Learning rate: 5e-5, 4e-5, 3e-5, 2e-5
  \item Number of epochs: 2, 3, 4
\end{itemize}

Here, a type of learning rate scheduler is linear,
Adam~\citep{kingma2017adammethodstochasticoptimization} is utilized for the optimizer, 
a number of weight decay is 0.01,
warmup steps is fixed to 500, 
a seed value is fixed to the same number through all evaluation experiments,
and the other training hyperparameters follow the default values of Training Arrguments library.

\subsubsection{Training Arguments for ALBERT and RoBERTa}
We fine-tune ALBERT and RoBERTa for GLUE downstream tasks.
The following hyperparameters are adopted in the experiments:

\begin{itemize}
  \item Batch size: 32
  \item Learning rate: 4e-5
  \item learning rate scheduler: linear
  \item Optimizer: Adam~\citep{kingma2017adammethodstochasticoptimization}
  \item warmup steps: 500
  \item number of weight decay: 0.01
\end{itemize}

This combination of hyperparameters was chosen
because it yields the best when evaluating BERT on the GLUE validation data,
which is explained on Appendix~\ref{app:glue_bert}.

A seed value is fixed to the same number through all evaluation experiments,
and the other training hyperparameters follow the default values of Training Arrguments library.

\section{SEAT score for \textit{Gender} bias}\label{app:gender}
In this section,
we show the SEAT results focusing specifically on the \textit{gender} bias.
The reason for showing results only for \textit{gender} bias
is that this bias is the most widely studied in the context of debiasing LMs.

It is to be noted that
the experimental setup for the debias evaluation
follows the same configuration as described in 
Section~\ref{sec:experiments_debias_eval}.

\subsection{Evaluation Metrics}
In addition to the bias measurement (Equation~\ref{eq:d}),
we show the permutation test for each dataset,
defined as follows:

\begin{equation}
  \label{eq:pvalue}
    p = \Pr \left[ s(X_i^*, Y_i^*, A, B) > s(X, Y, A, B) \right],
\end{equation}
where $(X_i, Y_i)$ is a subset of $X \cup Y$.

$s(X,Y,A,B)$ is obtained through the following formula:

  \begin{align}
  \label{eq:s-xyab}
    s&(X,Y,A,B) \\
    &= \sum_{x \in X} s(x,A,B) - \sum_{y \in Y} s(y,A,B).
  \end{align}

\subsection{Comparing Methods}
This section compares other approaches with the Bias Vector method.
These methods are selected based on the emperical study by \citet{meade-etal-2022-empirical}.

\subsubsection{Comparing Methods for Gender Bias}
This section explains the four existing methods
that have been used in \textit{gender} bias mitigation experiments.

\textbf{Counterfactual Data Augmentation (CDA)} \citep{zmigrod-etal-2019-counterfactual,dinan-etal-2020-queens,Webster2020,barikeri-etal-2021-redditbias}:
The CDA
involves creating data
by swapping biased words in text data
(such as \{she / he\} for the \textit{gender} bias).

\textbf{Dropout} \citep{Webster2020}:
The Dropout method
attempts to reduce bias by increasing the dropout parameters
that is originally used to mitigate the \textit{gender} bias.


\textbf{Iterative Nullspace Projection (INLP)} \citep{ravfogel-etal-2020-INLP}:
The INLP is a debiasing method
that uses a classifier to predict bias types (e.g., \textit{gender});
it then projects the embeddings into the null-space of that classifier
for eliminating information.
This process is iteratively applied to debias the embeddings of LM outputs.

\textbf{SentDebias} \citep{liang-etal-2020-towards}:
The SentDebias technique
extends the word embedding debiasing technique (Hard-Debias)
proposed by \citet{NIPS2016_a486cd07} to sentence embeddings.
SentDebias estimates a linear subspace of a specific bias
and removes the bias by projecting the sentence embeddings into this subspace.


\subsection{Results and Discussion}
\begin{figure}[t]
  \includegraphics[width=\columnwidth]{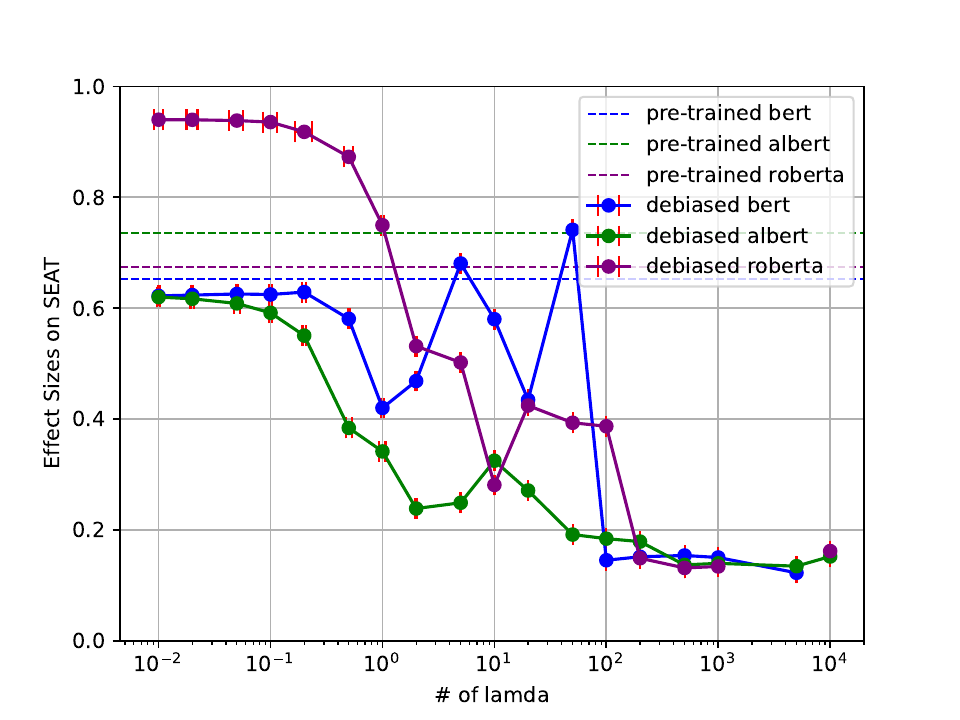}
  \caption{Effect sizes on gender bias tests in SEAT when varying the value of $\lambda$. The dashed lines indicate effect sizes on pre-trained LMs.}
  \label{fig:lambda_gender}
\end{figure}

The detail results on SEAT regarding \textit{gender} bias
are shown in Table~\ref{tab-gender} and Figure~\ref{fig:lambda_gender}.

It was confirmed that BV(\textit{all}, 1) yields
better than BV(\textit{gender}, 1).
Two reasons are considered
for why BV(\textit{gender}, 1) did not work sufficiently.
First, words indicating gender, such as \{she / he\},
likely appeared frequently in the pre-training corpus.
This high frequency made only a small difference
between pre-trained LMs and biased ones,
therefore,
the Bias Vector could not capture
enough \textit{gender} bias.
Second, the amount of data
used to continually train LMs toward \textit{gender} bias
was limited (996 instances).
This data limitation suggests that the data volume might be insufficient.

Furthermore,
it can be said that
BV(\textit{all}, 1)
debiased across all LMs,
and was sometimes competitive
with existing methods specialized in embedding spaces.
Additionally,
by adjusting $\lambda$ to 10 or 100,
BV(\textit{gender}, $\lambda$) results outperformed
the existing methods except for INLP on BERT.

\begin{table*}[t]
  \centering
  \scalebox{0.93}{
    \begin{tabular}{l|r r r r r r rr}
      \hline
      \multicolumn{1}{c|}{Methods} &	SEAT-6 & SEAT-6b & SEAT-7 & SEAT-7b & SEAT-8 & SEAT-8b & \multicolumn{2}{c}{Average ($\downarrow$)} \\ \hline\hline
      BERT & 0.932* & 0.090 & -0.124 & 0.937* & 0.783* & 0.858* && 0.621 \\
      ~w/ CDA & 0.846* & 0.186 & -0.278 & 1.342* & 0.831* & 0.849* & \cellcolor{red!15}\({\uparrow 0.101}\) & 0.722 \\
      ~w/ Dropout & 1.136* & 0.317 & 0.138 & 1.179* & 0.879* & 0.939 & \cellcolor{red!15}\({\uparrow 0.144}\) & 0.765 \\
      ~w/ INLP & 0.317 & -0.354 & -0.258 & 0.105 & 0.187 & -0.004 & \cellcolor{blue!15}\({\downarrow 0.417}\) & \textbf{0.204} \\
      ~w/ SentDebias & 0.350 & -0.298 & -0.626 & 0.458* & 0.413 & 0.462* & \cellcolor{blue!15}\({\downarrow 0.187}\) & 0.434 \\
      ~w/ BV(all, $1$) & 0.979* & 0.021* & -0.344 & 0.829* & 0.701* & 0.828* & \cellcolor{blue!15}\({\downarrow 0.004}\) & 0.617 \\
      ~w/ BV(gender, $1$) & 0.937 & 0.089 & -0.146 & 0.942 & 0.774 & 0.852 & \cellcolor{red!15}\({\uparrow 0.002}\) & 0.623\\
      ~w/ BV(gender, $10$) & 0.962* & 0.078 & -0.257 & 0.901* & 0.739* & 0.780* & \cellcolor{blue!15}\({\downarrow 0.002}\) & 0.619\\
      ~w/ BV(gender, $100$) & 0.760 & -0.060 & -0.107 & 0.482* & 0.188 & 0.266 & \cellcolor{blue!15}\({\downarrow 0.311}\) & 0.310\\
      \hline

      ALBERT & 0.637* & 0.151 & 0.487* & 0.956* & 0.683* & 0.823* && 0.623 \\
      ~w/ CDA & 1.040* & 0.170 & 0.830* & 1.287* & 1.212* & 1.179* & \cellcolor{red!15}\({\uparrow 0.330}\) & 0.953 \\
      ~w/ Dropout & 0.506* & 0.032 & 0.661* & 0.987* & 1.044* & 0.949* & \cellcolor{red!15}\({\uparrow 0.074}\) & 0.697 \\
      ~w/ INLP & 0.574* & -0.068 & -0.186 & 0.566* & 0.161 & 0.518* & \cellcolor{blue!15}\({\downarrow 0.278}\) & 0.345 \\
      ~w/ SentDebias & 0.490* & -0.026 & -0.032 & 0.489* & 0.431 & 0.647* & \cellcolor{blue!15}\({\downarrow 0.271}\) & 0.352 \\
      ~w/ BV(all, $1$) & 0.311 & 0.019 & 0.345 & 0.612* & 0.509 & 0.569 & \cellcolor{blue!15}\({\downarrow 0.229}\) & 0.394 \\
      ~w/ BV(gender, $1$) & 0.636* & 0.151 & 0.479* & 0.946 & 0.673* & 0.813 & \cellcolor{blue!15}\({\downarrow 0.007}\) & 0.616 \\
      ~w/ BV(gender, $10$) & 0.643* & 0.127 & 0.396* & 0.508* & 0.590* & 0.701* & \cellcolor{blue!15}\({\downarrow 0.129}\) & 0.494 \\
      ~w/ BV(gender, $100$) & -0.370 & -0.162 & 0.475* & 0.236 & 0.130 & 0.253 & \cellcolor{blue!15}\({\downarrow 0.441}\) & \textbf{0.182} \\
      \hline

      RoBERTa & 0.922* & 0.208 & 0.979* & 1.460* & 0.810* & 1.261* && 0.940 \\
      ~w/ CDA & 0.976* & 0.013 & 0.848* & 1.288* & 0.994* & 1.160* & \cellcolor{blue!15}\({\downarrow 0.060}\) & 0.880 \\
      ~w/ Dropout & 1.134* & 0.209 & 1.161* & 1.482* & 1.136* & 1.321* & \cellcolor{red!15}\({\uparrow 0.134}\) & 1.074 \\
      ~w/ INLP & 0.812* & 0.059 & 0.604* & 1.407* & 0.812* & 1.246* & \cellcolor{blue!15}\({\downarrow 0.117}\) & 0.823 \\
      ~w/ SentDebias & 0.755* & 0.068 & 0.869* & 1.372* & 0.774* & 1.239* & \cellcolor{blue!15}\({\downarrow 0.094}\) & 0.846 \\
      ~w/ BV(all, $1$) & 0.829* & 0.187 & 0.943* & 1.46* & 0.724* & 1.220* & \cellcolor{blue!15}\({\downarrow 0.046}\) & 0.894 \\
      ~w/ BV(gender, $1$) & 0.914* & 0.203* & 0.983* & 1.47* & 0.822 & 1.264* & \cellcolor{red!15}\({\uparrow 0.002}\) & 0.942 \\
      ~w/ BV(gender, $10$) & 0.845* & 0.153 & 0.905* & 1.515* & 0.908* & 1.273* & \cellcolor{blue!15}\({\downarrow 0.007}\) & 0.933 \\
      ~w/ BV(gender, $100$) & 0.517* & 0.041 & -0.366 & 1.173 & -0.144* & 0.842* & \cellcolor{blue!15}\({\downarrow 0.426}\) & \textbf{0.514} \\
      \hline
    \end{tabular}
  }
  \caption{\label{tab-gender}
    Effect sizes on SEAT with pre-trained or debiased LMs (BERT, ALBERT and RoBERTa) in gender bias tests.
    Average presents the mean of absolute effect sizes across all six gender tests for each LMs.
    Effect sizes closer to 0 suggest that LM representations are less biased.
    Statistically significant effect sizes with p-values lower than 0.01 are marked with *.
    All results of the existing methods are cited from \citet{meade-etal-2022-empirical}.
  }
\end{table*}

\section{Results in Each SEAT dataset}\label{app:seat_dev}
In this section,
we show the results for subset of SEAT dataset, SEAT-8 and SEAT-5b,
with means and standard deviations of effect sizes over ten seed values.

\begin{figure*}[t]
  \centering
  \begin{minipage}[b]{0.45\linewidth}
      \centering
      \includegraphics[width=1.0\linewidth]{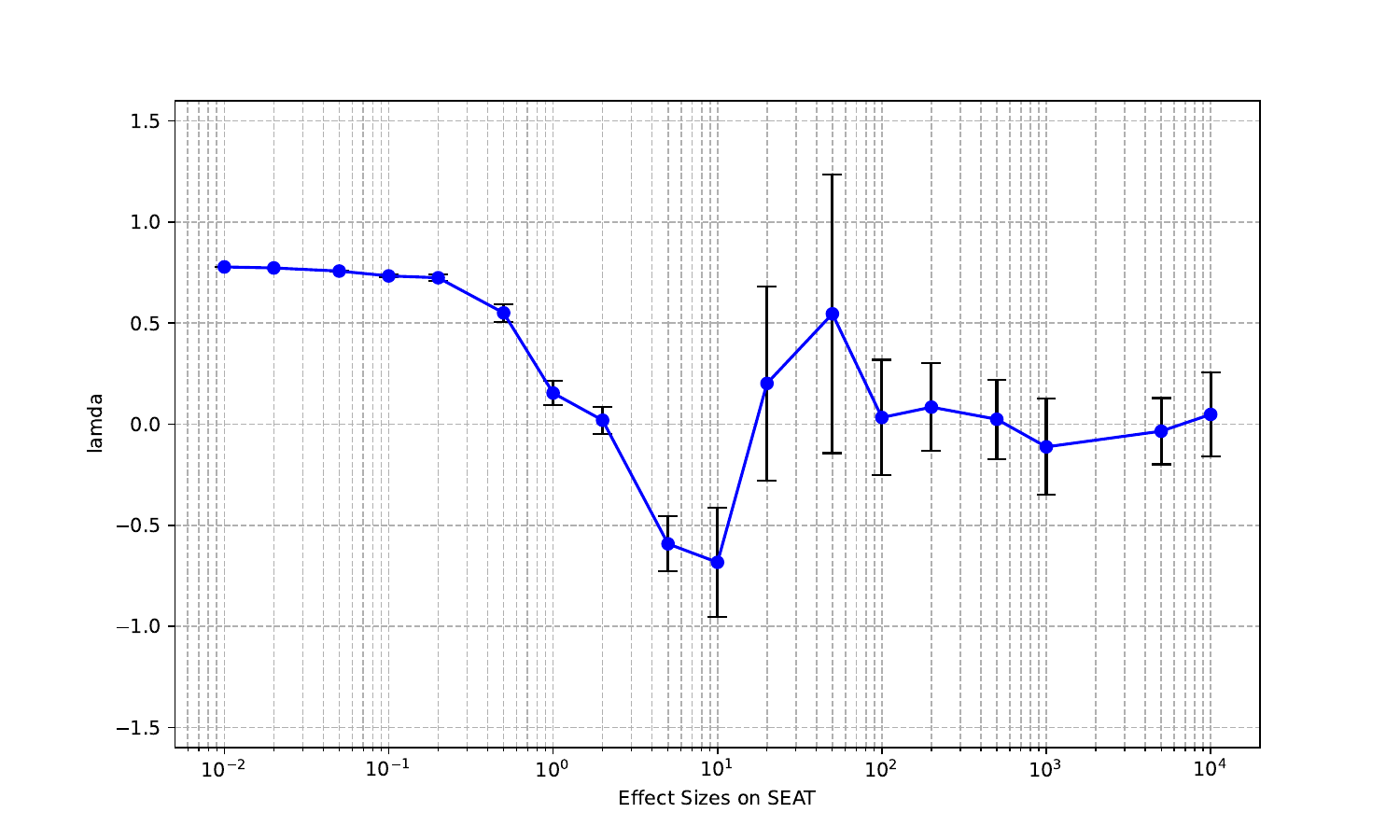}
      \caption{Means and standard deviations of effect sizes on SEAT-8 with debiased BERT.}
      \label{fig:bert_seat8}
  \end{minipage}
  \hfill
  \begin{minipage}[b]{0.45\linewidth}
      \centering
      \includegraphics[width=1.0\linewidth]{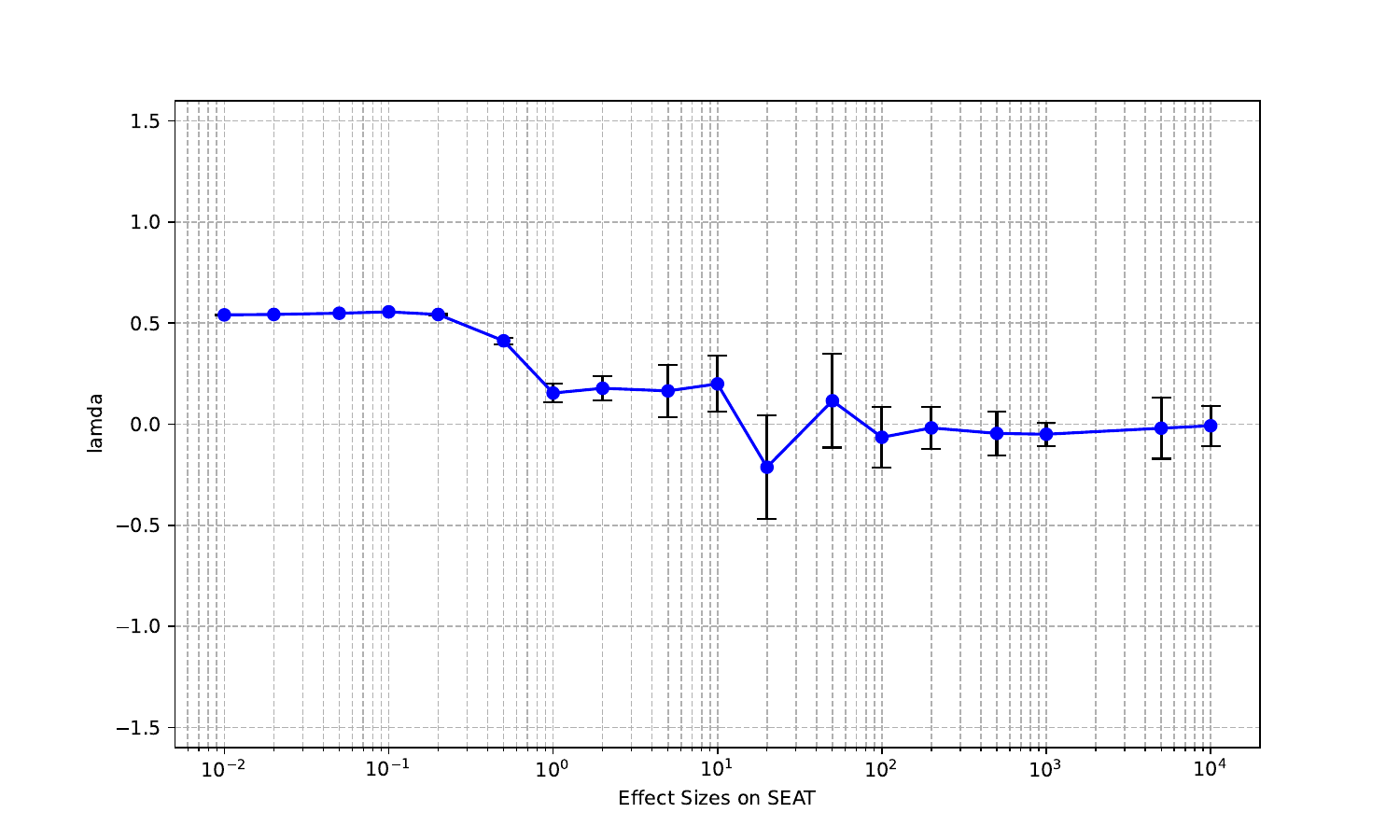}
      \caption{Means and standard deviations of effect sizes on SEAT-5b with debiased BERT.}
      \label{fig:bert_seat5b}
  \end{minipage}
  \vspace{0.5cm}

  \begin{minipage}[b]{0.45\linewidth}
      \centering
      \includegraphics[width=1.0\linewidth]{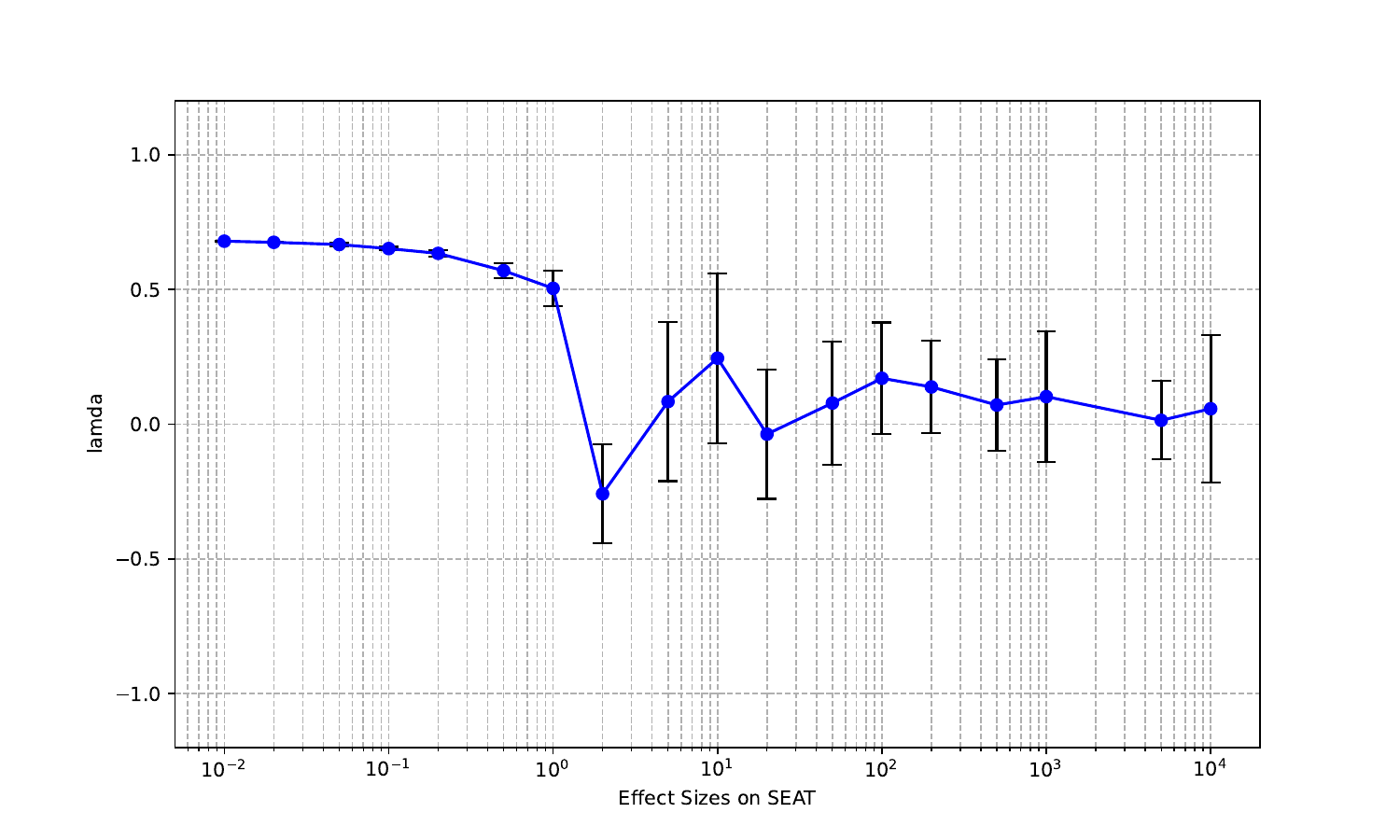}
      \caption{Means and standard deviations of effect sizes on SEAT-8 with debiased ALBERT.}
      \label{fig:albert_seat8}
  \end{minipage}
  \hfill
  \begin{minipage}[b]{0.45\linewidth}
      \centering
      \includegraphics[width=1.0\linewidth]{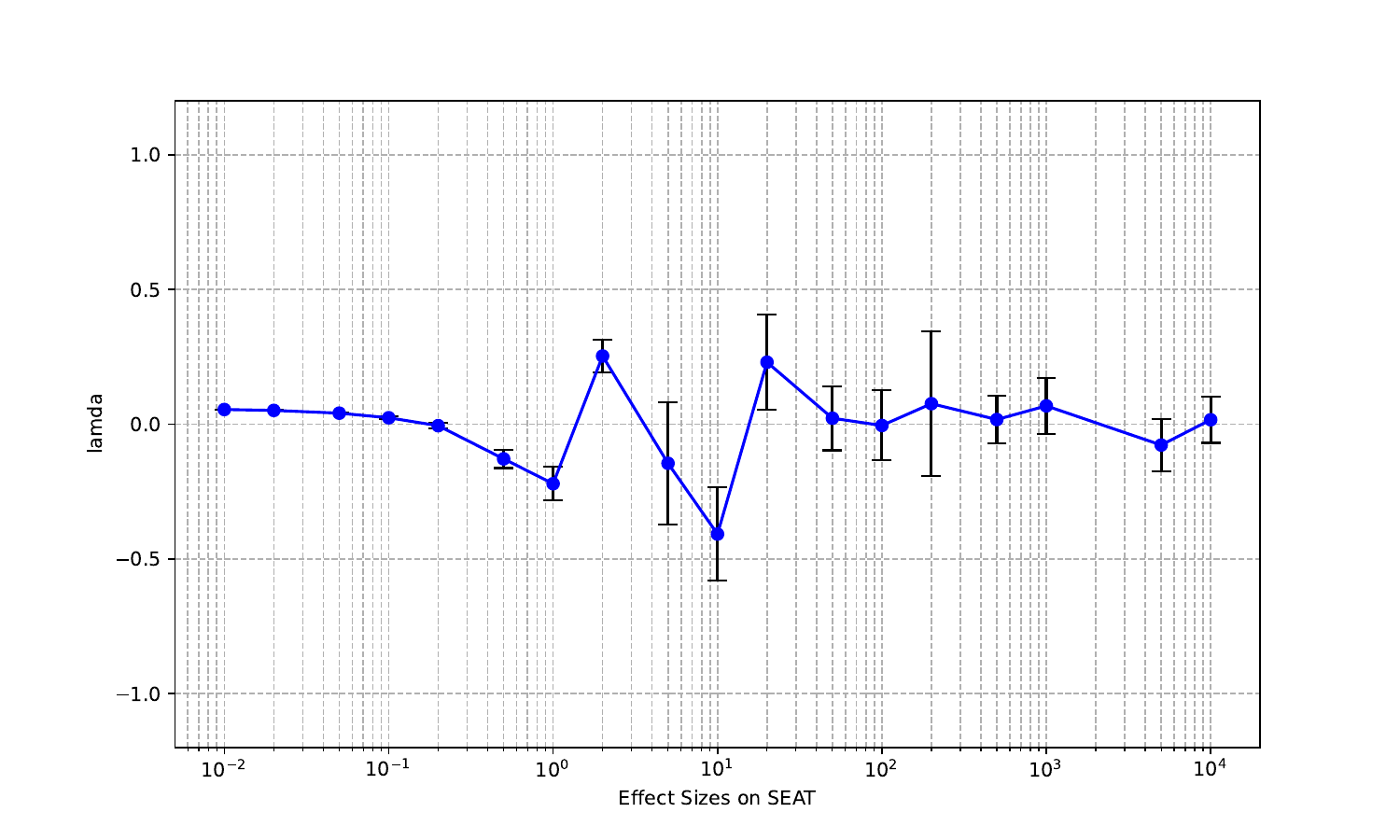}
      \caption{Means and standard deviations of effect sizes on SEAT-5b with debiased ALBERT.}
      \label{fig:albert_seat5b}
  \end{minipage}
  \vspace{0.5cm}

  \begin{minipage}[b]{0.45\linewidth}
      \centering
      \includegraphics[width=1.0\linewidth]{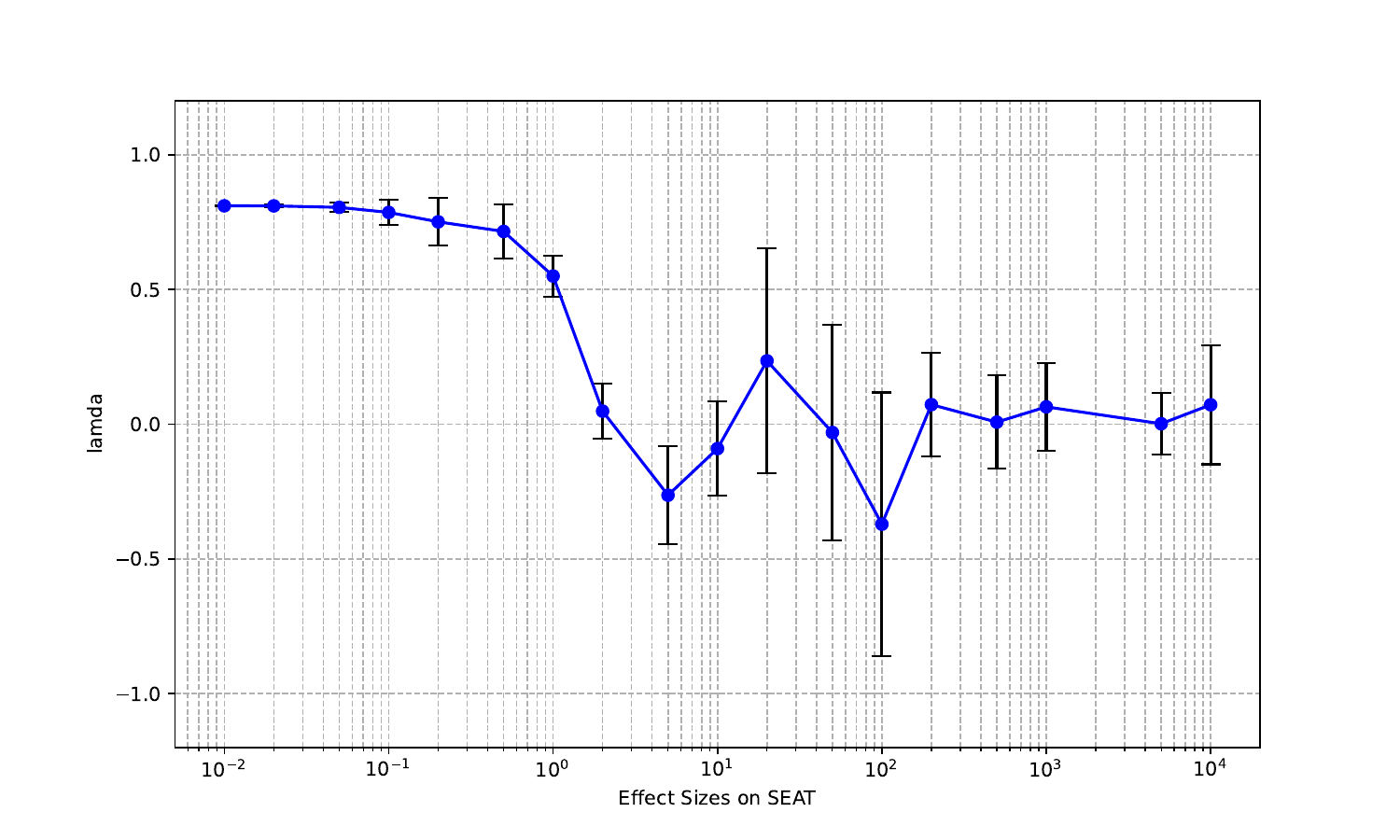}
      \caption{Means and standard deviations of effect sizes on SEAT-8 with debiased RoBERTa.}
      \label{fig:roberta_seat8}
  \end{minipage}
  \hfill
  \begin{minipage}[b]{0.45\linewidth}
      \centering
      \includegraphics[width=1.0\linewidth]{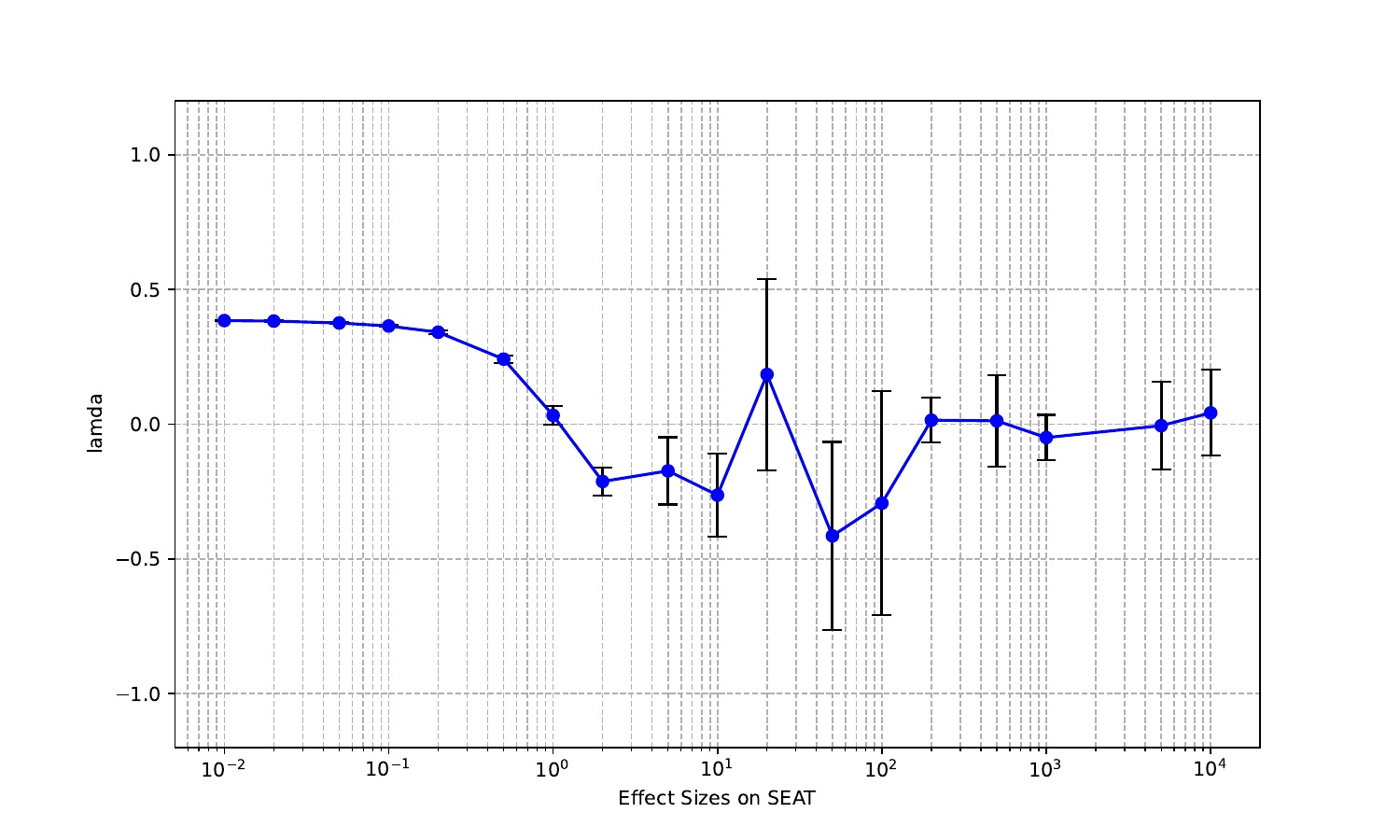}
      \caption{Means and standard deviations of effect sizes on SEAT-5b with debiased RoBERTa.}
      \label{fig:roberta_seat5b}
  \end{minipage}
\end{figure*}

We present the results of SEAT-8 dataset
in Figure~\ref{fig:bert_seat8} (BERT),
Figure~\ref{fig:albert_seat8} (ALBERT),
and Figure~\ref{fig:roberta_seat8} (RoBERTa).

The effect sizes with SEAT-5b dataset are described
in Figure~\ref{fig:bert_seat5b} (BERT),
Figure~\ref{fig:albert_seat5b} (ALBERT),
and Figure~\ref{fig:roberta_seat5b} (RoBERTa).

\end{document}